\documentclass[screen,nonacm]{acmart}
\title{What constitutes a Deep Fake?\\The blurry line between legitimate processing and manipulation under the EU AI Act} 

\author{Kristof Meding}
\affiliation{%
  \institution{Computational Law Lab, University of Tübingen}
  \city{Tübingen}
  \country{Germany}}
\email{Kristof.Meding@uni-tuebingen.de}

\author{Christoph Sorge}
\affiliation{%
  \institution{Chair of Legal Informatics, Saarland University}
  \city{Saarland Informatics Campus}
  \country{Germany}}
\email{christoph.sorge@uni-saarland.de}

\date{}

\usepackage[utf8]{inputenc}
\usepackage[T1]{fontenc}

\usepackage{graphicx,subcaption}

\begin{document}

\acmYear{2025}\copyrightyear{2025}
\acmConference[CSLAW '25]{Symposium on Computer Science and Law}{March 25--27, 2025}{München, Germany}
\acmBooktitle{Symposium on Computer Science and Law (CSLAW '25), March 25--27, 2025, München, Germany}
\acmDOI{10.1145/3709025.3712218}
\acmISBN{979-8-4007-1421-4/25/03}

\begin{abstract}
When does a digital image resemble reality? The relevance of this question increases as the generation of synthetic images---so-called deep fakes---becomes increasingly popular. Deep fakes have gained much attention for a number of reasons---among others, due to their potential to disrupt the political climate. In order to mitigate these threats, the EU AI Act implements specific transparency regulations for generating synthetic content or manipulating existing content. However, the distinction between real and synthetic images is---even from a computer vision perspective---far from trivial. We argue that the current definition of deep fakes in the AI Act and the corresponding obligations are not sufficiently specified to tackle the challenges posed by deep fakes. 
By analyzing the life cycle of a digital photo from the camera sensor to the digital editing features, we find that:
(1.) Deep fakes are ill-defined in the EU AI Act. The definition leaves too much scope for what a deep fake is.
(2.) It is unclear how editing functions like Google's ``best take'' feature can be considered as an exception to transparency obligations.
(3.) The exception for substantially edited images raises questions about what constitutes substantial editing of content and whether or not this editing must be perceptible by a natural person.

Our results demonstrate that complying with the current AI Act transparency obligations is difficult for providers and deployers. As a consequence of the unclear provisions, there is a risk that exceptions may be either too broad or too limited. We intend our analysis to foster the discussion on what constitutes a deep fake and to raise awareness about the pitfalls in the current AI Act transparency obligations.

\end{abstract}


\maketitle

\section{Introduction}
Generative AI tools that can generate photorealistic images or manipulate existing photos in a convincing manner are now widely available. The term ``deep fake'' has been coined, referring to images generated or manipulated using deep neural networks. Actual deep fakes and allegations of their use play an increasingly important role in political debates and election campaigns as well as in pornographic content \cite{becker_generative_2024,kristof2024online}. Given the potential impact of manipulated images, regulators increasingly identify a need to regulate the use of such images. Countries that have already started initiatives to regulate deep fakes include the US, China, the UK, and South Africa \cite{ramluckan_deepfakes_2024, murray_generative_2024}.

\begin{figure}
\centering
\begin{subfigure}{.51\columnwidth}
  \centering
  \includegraphics[width=.8\columnwidth,height=.79\columnwidth]{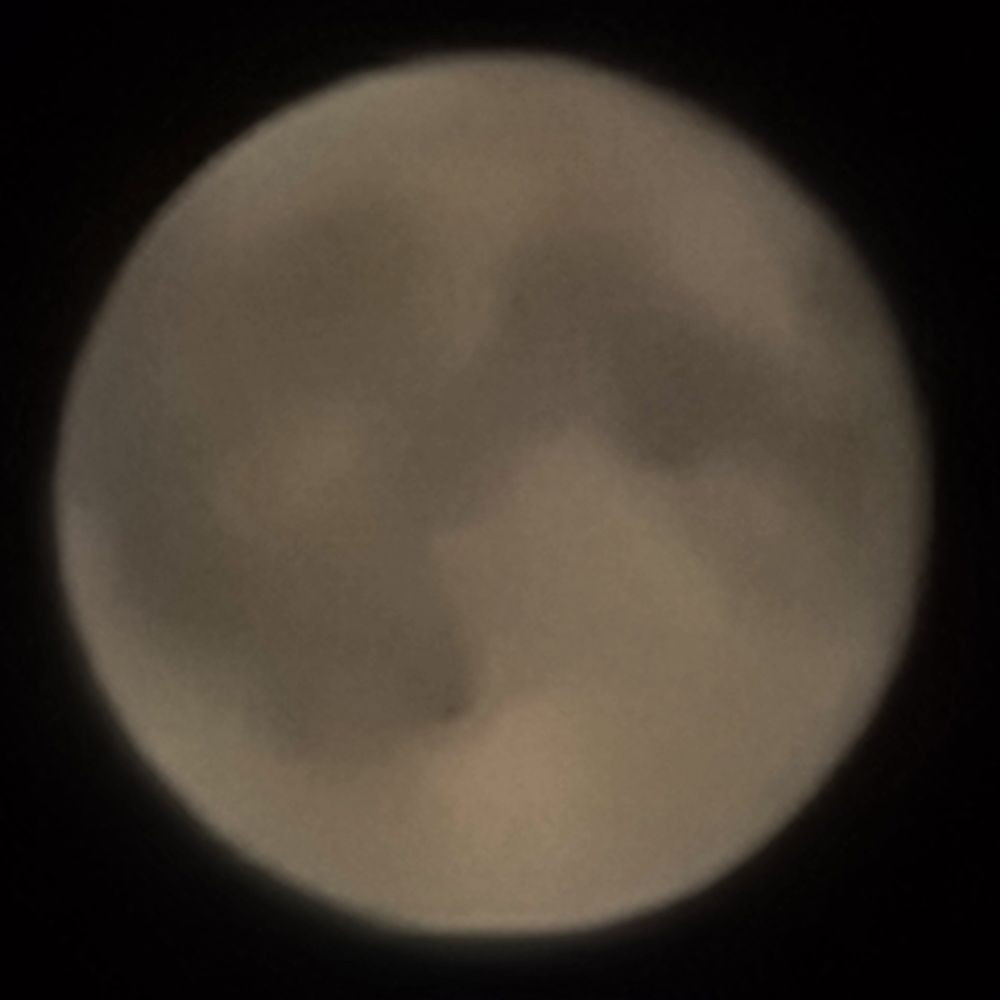}
            \vspace{1em}
  \caption{w/o Scene Optimizer}
  \label{fig:sub1}
  \Description{A very blurry image of the full moon.}
\end{subfigure}%
\begin{subfigure}{.5\columnwidth}
  \centering
  \includegraphics[width=.8\columnwidth,height=.8\columnwidth]{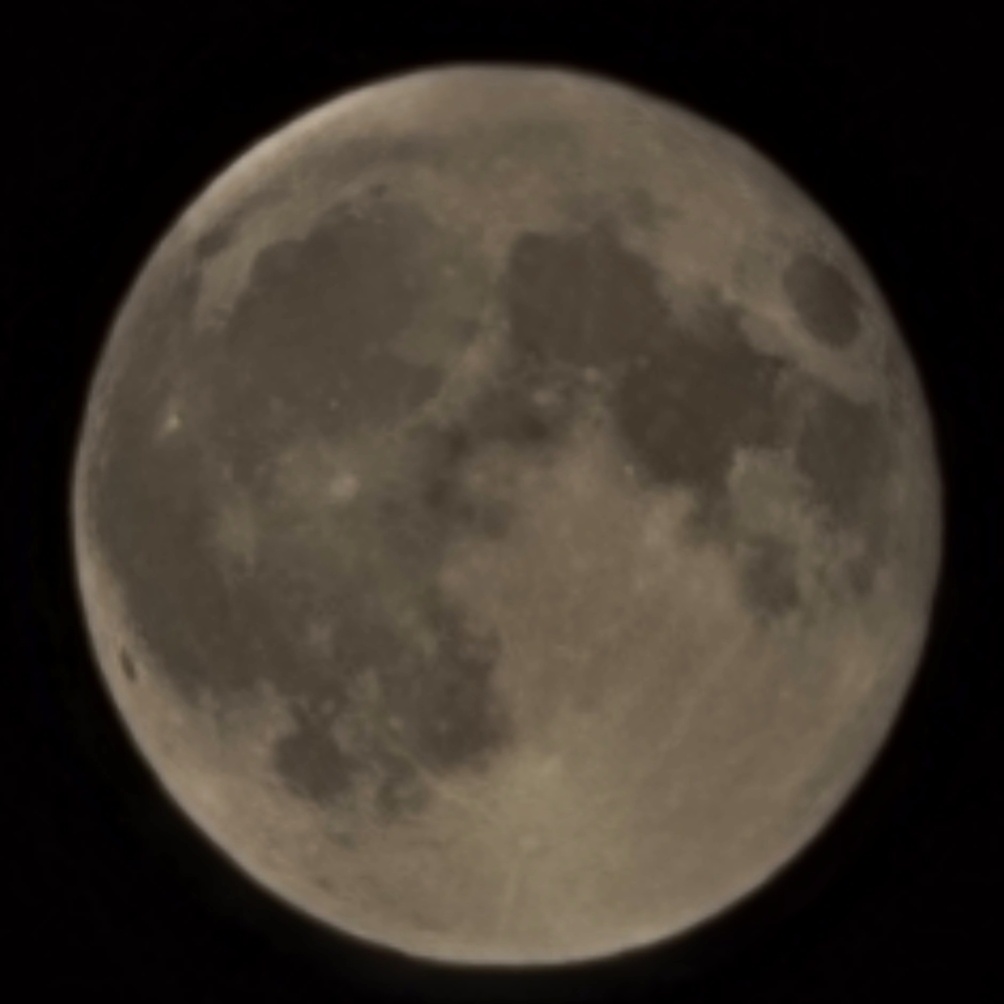}
            \vspace{1em}
  \caption{With Scene Optimizer}
  \label{fig:sub2}
    \Description{An image of the full moon with significantly more detail than the left one.}
\end{subfigure}
\caption{Details of moon photographs taken with a Samsung Galaxy S20 5G smartphone by Thomas Wagner.}
\label{fig:moon}
\end{figure}

The European Union (EU)'s AI Act includes a transparency obligation for AI systems generating deep fakes and other synthetic content. The AI Act's definition relates to ``manipulated'' content, which raises the question of how to differentiate between manipulation and permissible editing. Photos have never been an objective depiction of reality. In the context of photojournalism, an author has called for framing ``the production and use of reality images as `mediated communication' rather than an `objective truth''' \cite{newton2013burden}. The photographer's location and the focal length used, the depth of field and the lighting can create completely different visual impressions of the same object, even in analog photography and without post-processing. As seen in advertising photography, this is often done with manipulative intent. Digital post-processing opens up additional possibilities, such as adaptations of white balance and the color palette or the removal of spots caused by dust on the image sensor. In recent years, an increasing range of AI-based post-processing tools has become available to amateur photographers and smartphone users. The camera app pre-installed on some Samsung phones automatically replaces depictions of the moon by an improved and adapted image (see Fig.~\ref{fig:moon}). Google's recently presented ``Best Take'' feature allows swapping faces, enabling users to ensure that all persons depicted in a group photo are shown with a smile and their eyes open. This paper aims to open the discussion on where to draw the line between legitimate processing and manipulative deep fakes. It focuses on the interpretation of the European AI Act, but also raises fundamental questions.

 While previous work has discussed how the AI Act impacts image forensics in regard to deep fake detection \cite{lorch_compliance_2022}, we focus on the generation of deep fakes either from scratch or through the manipulation of (existing/real) content, and the corresponding transparency regulations. The literature has already examined whether AI systems that generate deep fakes can be considered high-risk systems (Article 6 AI Act\footnote{Regulation (EU) 2024/1689.} (AIA)) or general-purpose AI models (Article 3(66) AIA)) \cite{block2024critical}. We do not include this discussion in this paper. 

The remainder of this paper is structured as follows:
Section \ref{sec::Sec2} presents common image processing techniques, both traditional and with the support of AI. In Section \ref{sec:Regulation}, we provide the background on the regulation of deep fakes in the European AI Act. Based on the two previous sections, we discuss the challenges and edge cases of deep fake regulation in Section \ref{sec:Challenges}, and conclude with an outlook in Section \ref{sec:outlook}.

\section{Image processing -- Lifecycle of a digital photo}
\label{sec::Sec2}
\subsection{Traditional}
\label{sec::post-processing-traditional}
Digital photos must always be processed to be intelligible to a viewer. The following steps are often performed automatically (e.g., within the Android camera subsystem), and without the user necessarily being aware of them \cite{minhaz20android}:
\begin{figure*}
\centering
\begin{subfigure}{0.5\textwidth}
  \centering
  \includegraphics[width=0.96\columnwidth]{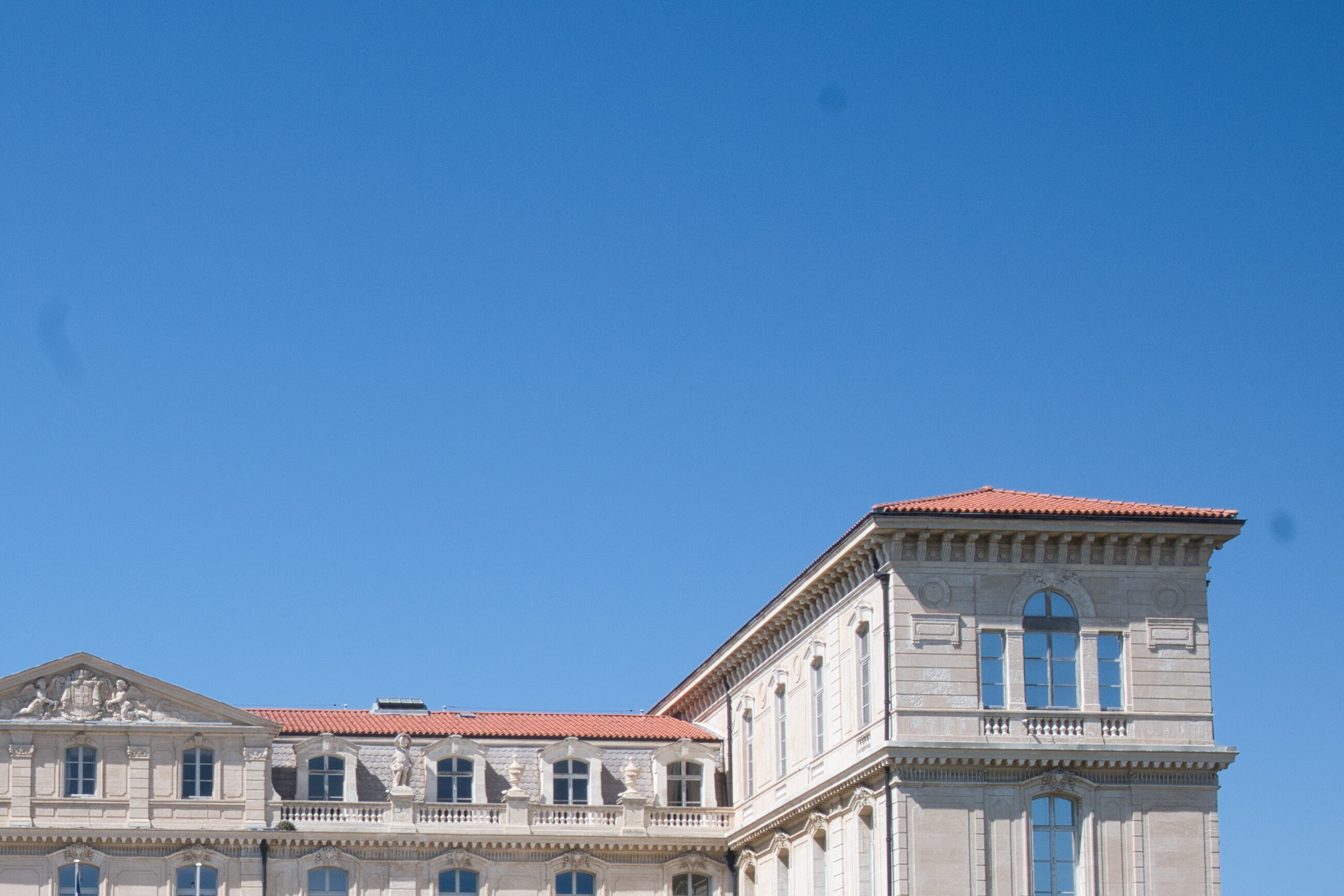}
  \caption{Original unmodified image}
  \label{fig:dustsub1}
    \Description{Parts of a building and a blue, cloudless sky with some darker spots (specks of dust).}
\end{subfigure}%
\begin{subfigure}{0.5\textwidth}
  \centering
  \includegraphics[width=0.96\columnwidth]{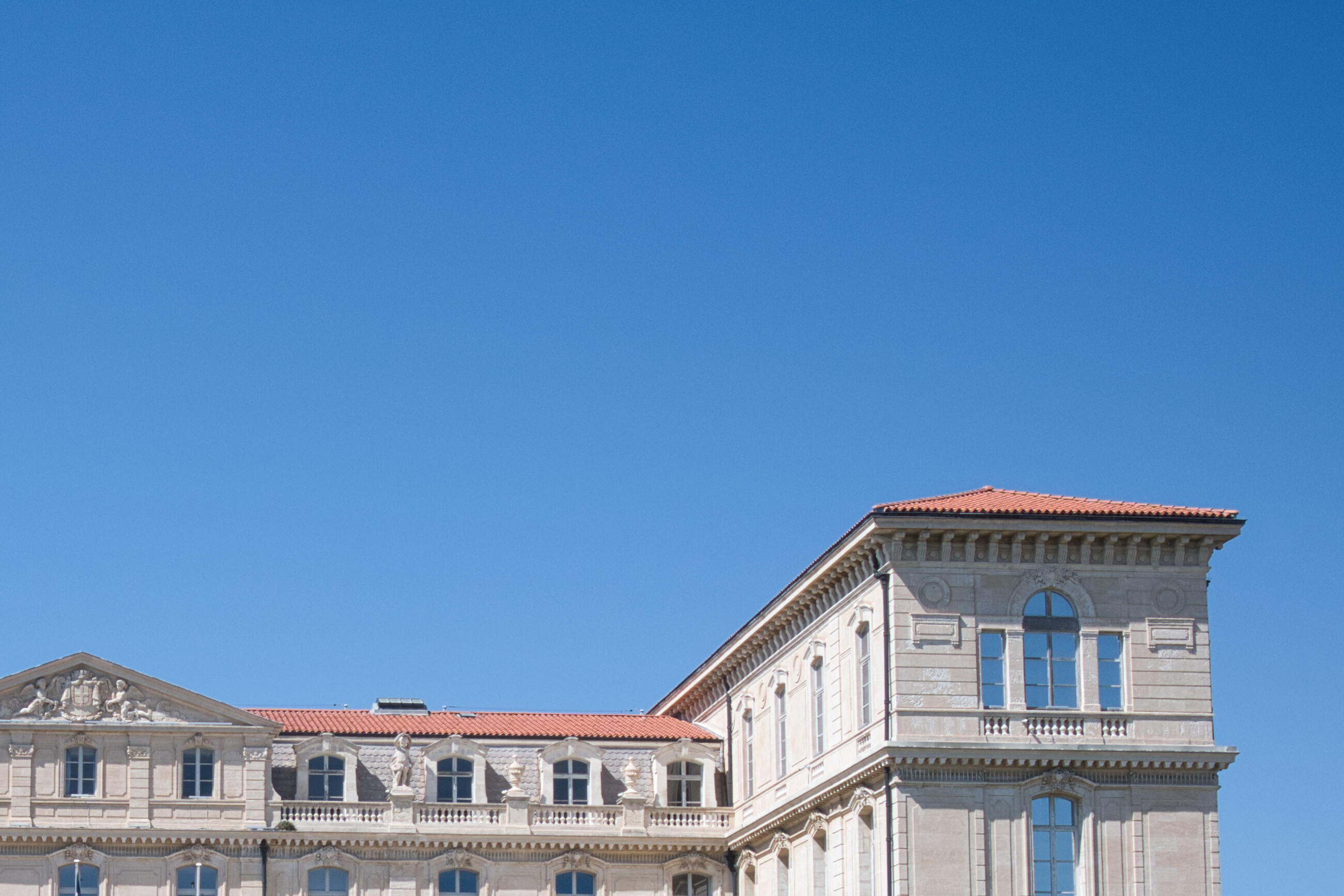}
  \caption{Version with removed dust}
  \label{fig:dustsub2}
      \Description{Parts of a building and a blue, cloudless sky.}
\end{subfigure}
\caption{Demonstration of dust removal (detail of a photo, dust removal performed with Luminar Neo without human intervention). Image best viewed in colour. }
\label{fig:dust}
\end{figure*}

\begin{itemize}
    \item Image sensors usually sense only one of the RGB colors per pixel, so the actual color of each pixel has to be interpolated.
    \item Faulty pixels are difficult to avoid during the production of image sensors, so they have to be detected and addressed.
    \item Sensor noise (random fluctuations of brightness and color), which is most problematic in low light situations, can be removed, potentially at the cost of losing some image detail.
    \item Camera lenses introduce additional image errors, such as vignetting (reduction of brightness and saturation towards the corners of the image) and distortion, which are often fixed automatically as well.
    \item Differences in lighting (e.g. daylight vs. artificial light) change the color of an object; to match human perception, which accounts for these changes, a correction needs to be applied. Colors are also corrected to account for the dynamic range of output devices (tone mapping).
    \item The image contrast around edges may be increased, so the image appears sharper.
\end{itemize}
Instead of performing these steps with the camera, professional photographers often choose to work on a mostly uncorrected (raw) image to get more control over the process. For instance, they may want to correct distortions caused by the camera perspective in addition to those introduced by the lens or adapt colors as an artistic choice instead of just correcting for lighting. Numerous additional filters and editing tools are available and commonly used. These include the removal of spots caused by dust on the image sensor (a common issue on cameras with interchangeable lenses), as well as sharpening and softening effects.

Some image processing tools use more than one photo as an input. Overlapping images can be combined into panoramic shots. Faint objects in the night sky are made visible by combining multiple exposures of the same region. Photos taken with different exposure times are merged to create an image with a higher dynamic range (HDR).

All these processing steps influence the perception of an image by a human viewer but are common or even necessary.

\subsection{AI-supported post-processing}
\label{sec::post-processing-ai}
Photographers use AI-based post-processing to an increasing extent: Although marketing claims about AI use are not a reliable indicator for a tool to be classified as AI, available tools provide functions that cannot be plausibly achieved without the use of AI.

\begin{figure*}[h!]
\centering
\begin{subfigure}{0.3\textwidth}
  \centering
  \includegraphics[width=0.96\columnwidth]{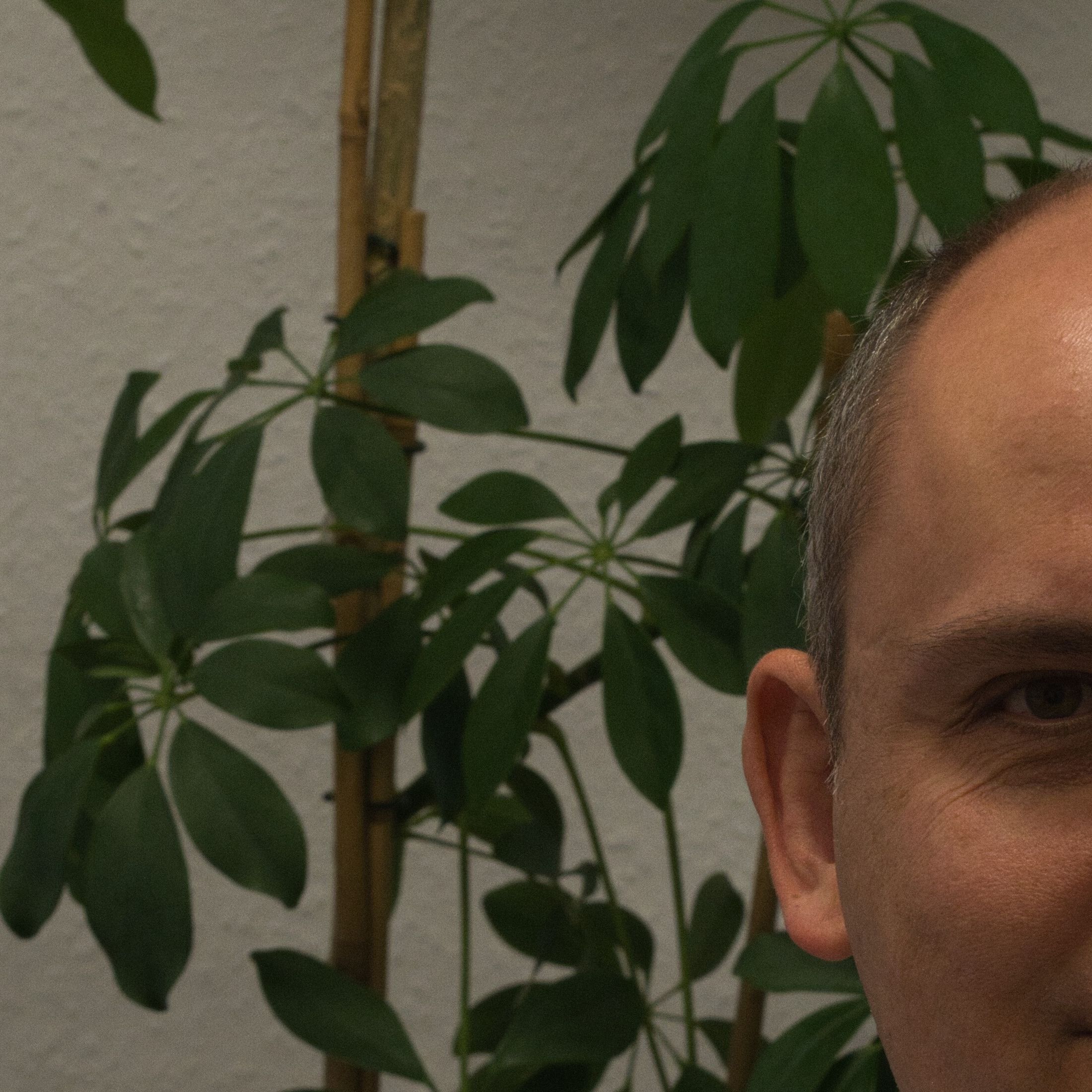}
  \caption{Original image, small aperture (f/14)}
  \label{fig:cssub1}
      \Description{A partial face in the foreground and a plant in the background. Both the face and the plant are sharply in focus.}
        \vspace{1em}
\end{subfigure}%
\begin{subfigure}{0.3\textwidth}
  \centering
  \includegraphics[width=0.97\columnwidth]{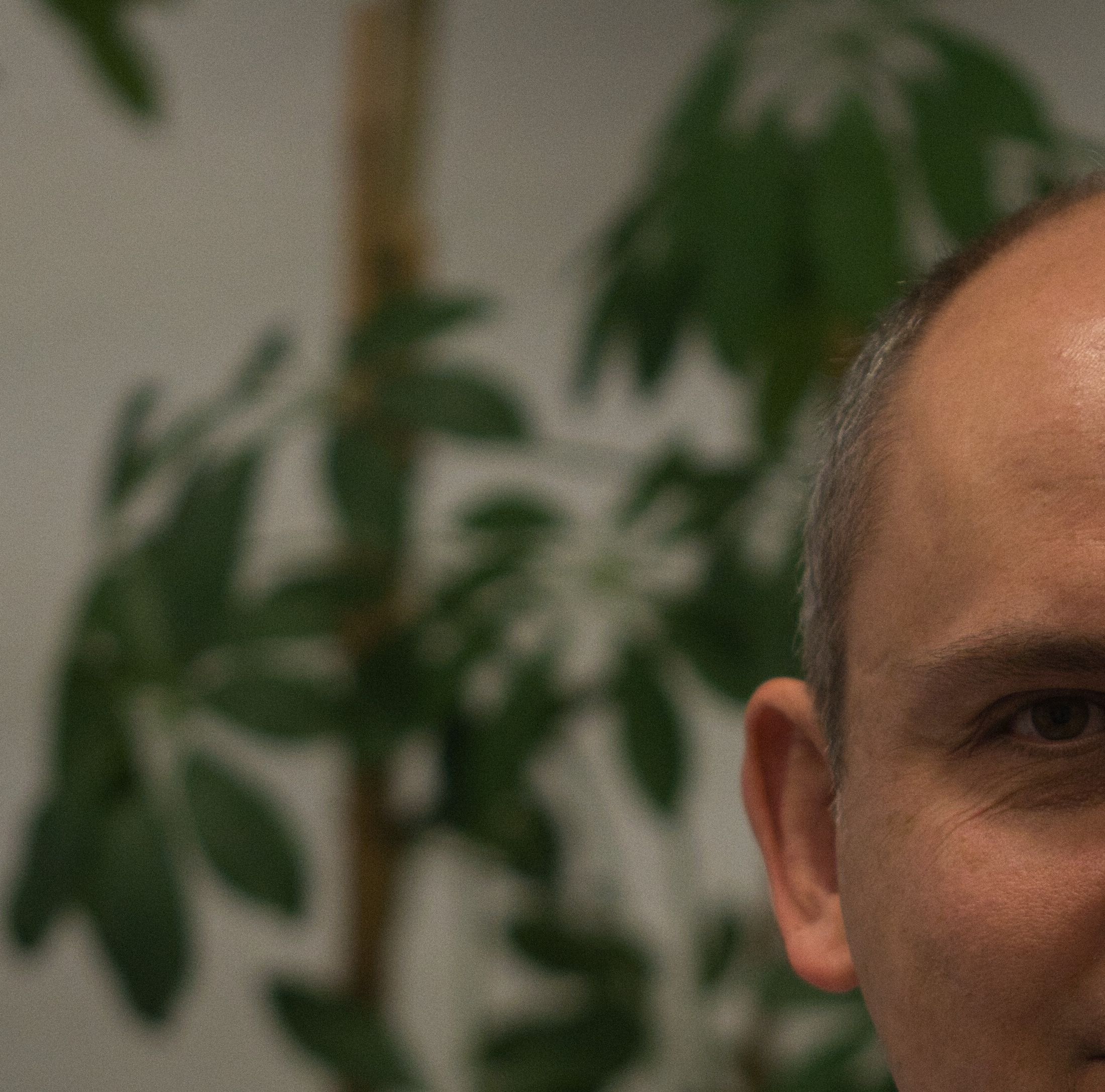}
  \caption{Original image, large aperture (f/3)}
  \label{fig:cssub2}
        \Description{A partial face in the foreground and a plant in the background. The face is sharply in focus, while the plant is blurry.}
          \vspace{1em}
\end{subfigure}
\begin{subfigure}{0.3\textwidth}
  \centering
  \includegraphics[width=0.96\columnwidth]{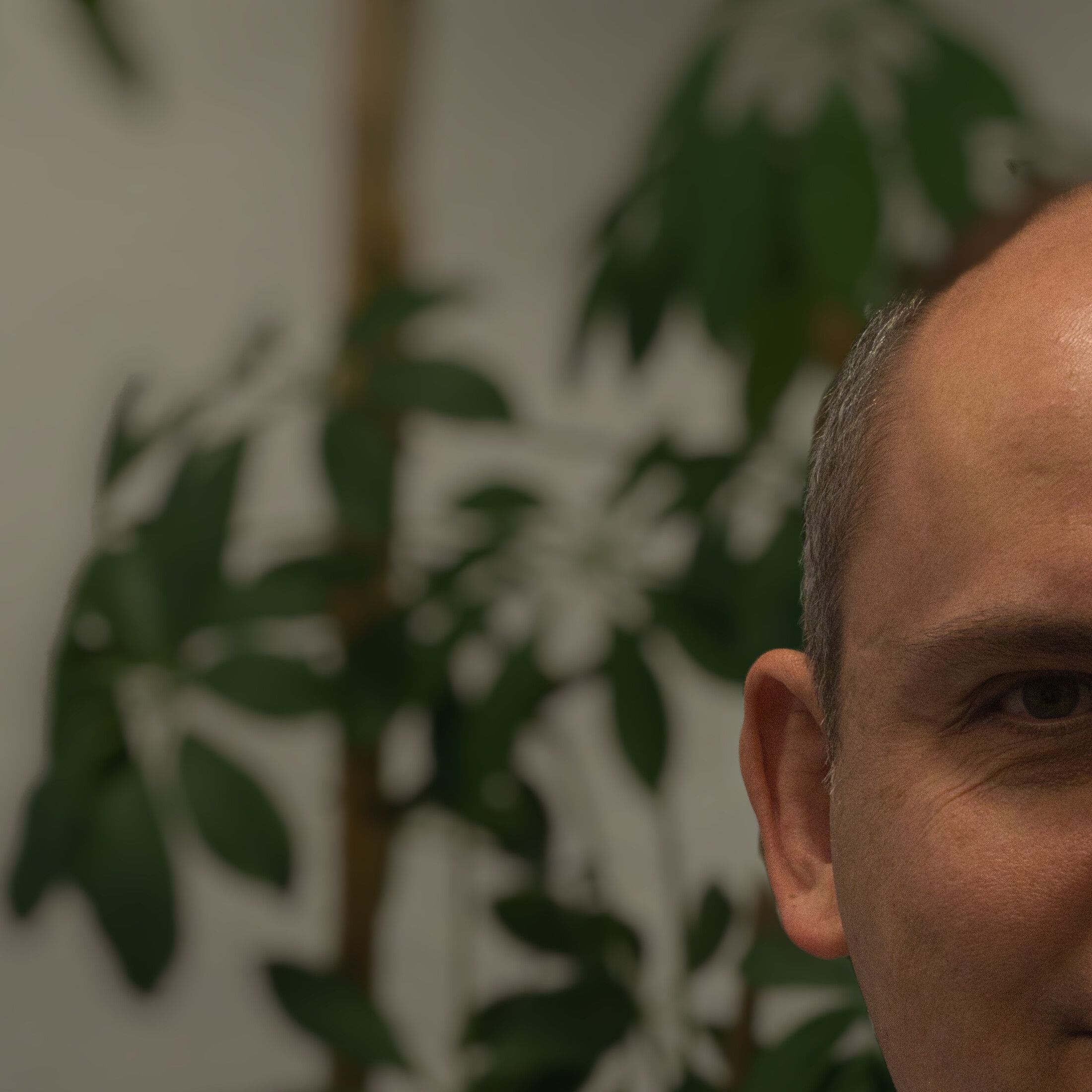}
  \caption{Processed version of the leftmost image}
  \label{fig:cssub3}
          \Description{A partial face in the foreground and a plant in the background. The face is sharply in focus, while the plant is blurry.}
\end{subfigure}
\caption{Demonstration of background blurring with Luminar Neo's ``Portrait Bokeh'' tool, showing details of an image}
\label{fig:blur}
\end{figure*}

Some of these functions replace processing that was previously available but required human intervention to a greater extent. The removal of image sensor dust commonly required users to mark the spots to be removed, but can now be performed automatically (see Fig.~\ref{fig:dust}). Similarly, power lines can be easily removed from an image as well\footnote{See, for instance, \url{https://skylum.com/en/how-to/how-to-remove-dust-spots-power-lines-automatically}.}. In order to blur the background in post-processing (which achieves a similar effect to choosing a wide aperture during the photo shoot, see Fig.~\ref{fig:blur}), photographers previously needed to mark the foreground; nowadays, this task can often be performed without intervention.

Other functions are only realistically possible due to the use of AI. Adobe Photoshop includes an ``AI Image Extender'' feature that allows to ``magically add more background''\footnote{\url{https://www.adobe.com/products/photoshop/generative-expand.html}.} using generative AI. Luminar Neo's BodyAI tool features a slider to ``slim or add volume to a subject’s torso''\footnote{\url{https://userguide.skylum.com/hc/en-us/articles/11542260006930-Working-with-the-Portrait-Tools}.}.

AI-based post-processing is commonly available on modern smartphones as well. Thus, photo-editing options are easily accessible to many users, and they might not even be aware of the functionality. 

\captionsetup{labelsep=space,justification=justified,singlelinecheck=off}

On March 10, 2023, Reddit user \emph{ibreakphotos} reported that ``Samsung `space zoom' moon shots are fake''\footnote{\url{https://www.reddit.com/r/Android/comments/11nzrb0/samsung_space_zoom_moon_shots_are_fake_and_here/}}. The post provides compelling evidence that the Scene Optimizer feature on at least some Samsung smartphones adds the texture of the moon to photos in which the moon is recognized, improving the image quality beyond the capabilities of current smartphone cameras. Within a few days, Samsung confirmed the use of deep learning to improve moon photos.\footnote{\url{https://www.samsungmobilepress.com/feature-stories/how-samsung-galaxy-cameras-combine-super-resolution-technologies-with-ai-technology-to-produce-high-quality-images-of-the-moon/}} Sample shots of the moon are depicted in Fig.~\ref{fig:moon}.

Google has implemented a feature called ``Best Take'' on Pixel smartphones, which is focused on improving pictures of persons (individuals and groups).\footnote{\url{https://store.google.com/intl/en/ideas/articles/pixel-best-take/}} The core idea of that feature has been described in a research paper \cite{rajvi12besttake}. A series of photos is taken instead of a single one. The software then analyzes facial expressions and selects the best image. Moreover, in a group photo, individual faces are seamlessly replaced by better versions (e.g., from a different take of the same photo series in which the respective person's eyes are open). \\

All processing steps mentioned above---being AI- or non-AI-based---pose challenges for deep fake regulation. 

\section{Regulation of deep fakes in the European AI Act}
\label{sec:Regulation}
Article 3(60) AIA defines deep fakes, and Article 50 AIA states resulting transparency obligations. Recitals 132 AIA to 137 AIA include further information on the interpretation of these transparency obligations. In the following sections, we discuss the core definition of deep fakes before presenting the resulting obligations. 

\subsection{Definition of deep fakes in the AI Act}
According to Article 3(60) AIA,
\begin{quotation}{‘deep fake’ means AI-generated or manipulated image, audio or video content that resembles existing persons, objects, places, entities or events and would falsely appear to a person to be authentic or truthful.}
\end{quotation}

According to the definition, four conditions need to be met in order for content to be considered a deep fake. First, the content needs to be generated or manipulated. Generation can be understood as (prompt-based) content creation \cite{gils_detailed_2024}. In contrast,  manipulation refers to given input data---for example, an already taken image---being altered. As discussed in Section \ref{sec::Sec2}, an image taken by a camera never shows the \emph(true) reality but instead is a result of the image processing chain as well as settings chosen by the photographer. For image content, defining the difference between traditional image processing steps and novel deep fake manipulation is problematic. We get back to this point in Section \ref{sec::ImageChainDeepFake}.

Second, image, audio, or video content is needed for a deep fake. Other content, such as text, is not considered deep fakes in the AI Act\footnote{There are, however, transparency obligations in Article 50 AIA, which also covers synthetic text content. We discuss this point in Section \ref{sec::Article50}.}.

Third, the content needs to resemble existing objects, places, entities, or events to be considered a deep fake. Thus, a link to the real world is necessary. Unicorns or monsters, which only give a photorealistic impression, are not considered deep fakes \cite{becker_generative_2024}. Recital 134 AIA includes a slightly modified version of the definition of a deep fake:
\begin{quotation} [\ldots] AI system to generate or manipulate image, audio, or video content that \textit{appreciably} resembles existing persons, objects, places, entities, or events and would falsely appear to a person to be authentic or truthful (deep fakes)[\ldots]
\end{quotation}
Adding the word \textit{appreciably} might slightly change the meaning of the definition, emphasizing the resemblance to existing objects~\cite{gils_detailed_2024}. The question arises ``whether the requirement of resemblance demands identicality or similarity to the existing subjects.'' \cite{gils_policy_2024}. For example, in the case of a face, the question is whether the face has to belong to a real person or whether the face has to be sufficiently similar to real persons. In this context, it has been argued that the difference between the definitions only shows the lack of regulatory precision and has no practical influence \cite{labuz_deep_2024}.

Finally, the content must appear to a person to be authentic or truthful. Thus, the impression to a natural person is important. Generated or manipulated content (such as adversarial perturbed images \cite{szegedy2013intriguing}) that is only classified as an existing person by an algorithm is not considered a deep fake. It has been argued that the requirement of an impression to \emph{a} person compared to an average person considerably broadens the scope \cite{block2024critical}. For example, some images might be perceived as authentic by children but not by adults \cite{block2024critical}. It is unclear which background knowledge and abilities of a person must be taken into consideration when classifying content as a deep fake. Moreover, available editing options in modern image processing software challenge the definition of a deep fake, see Section \ref{sec::ImageChainDeepFake}. 

Deep fakes are not only regulated in the EU AI Act, but also in other legislative acts of the EU \cite{labuz_regulating_2023}. Specifically, Article 35(1)(k) of the Digital Services Act (DSA) contains a similar wording to the AI Act's definition of deep fakes, which might pose challenges for the interplay between DSA and AI act.\footnote{See the cited paper for more information.} However, while previous versions of the AI act had different definitions of deep fakes, now at least a basic level of consistency between the AI act and other regulations such as the DSA has been reached \cite{labuz_deep_2024}. This basic level of consistency allows for partially consistent obligations between the different legislative acts.

\subsection{Transparency Obligations}
\label{sec::Article50}
Article 50 AIA of the AI Act includes the main transparency regulations and obligations concerning synthetic content and deep fakes. These obligations apply regardless of whether the underlying system is a high-risk system or general-purpose AI, as noted in recital 132.

Article 50(2) AIA  demands that 
\begin{quotation}
Providers of AI systems, including general-purpose AI systems, generating synthetic audio, image, video or text content, shall ensure that the outputs of the AI system are marked in a machine-readable format and detectable as artificially generated or manipulated. 
\end{quotation}
The first observation is that Article 50(2) AIA only covers providers (see Article 3(3) AIA for a definition of providers). In contrast, deployers (see Article 3(4) AIA)) are governed by Article 50(4)AIA.

Second, Article 50(2) AIA mentions only the generation of synthetic content. Here, the question arises whether the manipulation of existing content is also covered by the provision in Article 50(2) AIA\cite{gils_detailed_2024}. There are strong reasons to believe it is. In the latter part of the first sentence, Article 50(2) AIA mentions the obligation to mark the synthetic content as generated or manipulated. Recital 133 AIA explicitly also mentions the manipulation of existing inputs. Thus, we assume here that the generation and manipulation of input are covered by Article 50(2) AIA.

In contrast to the core deep fake regulation in Article 50(4) AIA, here, text content is also listed. Thus, the obligation of Article 50(2) AIA includes deep fakes but also extends beyond core deep fake regulation.

The main obligation is to mark the synthetic content in a machine-readable format to ensure that the content is detectable as artificially generated or manipulated. On the technical side, watermarking synthetic outputs is far from a trivial task \cite{gils_detailed_2024}, although Recital 133 ALA explicitly mentions some techniques. It remains to be seen whether the community will invest some joint effort to develop global standards \cite{romero_moreno_generative_nodate}.

Article 50(2) AIA does not explicitly mention that the marking of the synthetic output has to be detectable by human viewers\footnote{We specifically cover image content in our paper.} per se. Thus, it remains unclear whether there shall be a human-visible declaration of the synthetic content \cite{gils_detailed_2024}. However, Article 50(5)  AIA states that the information referred to in Article 50(2) shall be provided to natural persons in a clear and distinguishable manner. Marking content only in a machine-readable way without visibility to humans would be deceptive. So in practice, it does not matter whether Article 50(2) AIA alone or in conjunction with Article 50(5) AIA requires some human-visible marking. Additionally, providers of (large) social media platforms already have an obligation to label deep fakes through Article 35(1)(k) DSA \cite{block2024critical}. 

In contrast to the aforementioned minor issues in the transparency obligation, two exceptions in Article 50(2) AIA pose a major difficulty. Article 50(2) AIA states that
\begin{quotation}
This obligation [of Article 50(2) AIA] shall not apply to the extent the AI systems perform an assistive function for standard editing or do not substantially alter the input data provided by the deployer[\ldots]
\end{quotation}
So the first exception targets systems that perform assistive functions for standard editing tasks. However, the intention of the legislator remains unclear \cite{block2024critical}. There is no legal definition or a clarification in the recitals for either of the terms ``assistive function'' and ``standard editing''. What characterizes standard editing given the existence of modern AI-based image editing tools, such as Samsung's scene optimizer or Google's Best Take? The threshold for an assistive function or a standard editing tool is not fixed.

The second exception targets cases in which the input data or the semantics of the image are not substantially altered. The fundamental question here is: When is an image substantially altered? Are changes in the pixel-space or visible/perceptible alterations required to assume a ``substantial alteration''? 

From our point of view, both exceptions\footnote{There is also a third exception for law enforcement purposes which we do not consider in our paper.} pose a serious challenge to existing image editing tools on smartphones and computers. We discuss this issue in depth in Section \ref{sec:Challenges}.\\

The other main transparency obligation regarding deep fakes is given in Article 50(4) AIA subparagraph 1. Here, deep fakes are specifically mentioned:
\begin{quotation}
  Deployers of an AI system that generates or manipulates image, audio or video content constituting a deep fake, shall disclose that the content has been artificially generated or manipulated. 
\end{quotation}
In contrast to Article 50(2) AIA, this obligation only targets the deployers of an AI system. Furthermore, only deployers of systems processing image, audio, or video are regulated. Text systems, on the other hand, are not considered as deep fakes under the AI Act, and are regulated in Article 50(4) AIA subparagraph 2. It is unclear whether Article 50(4) AIA subparagraph 1 requires the AI content itself to be labeled---or whether labeling the system, for example, through a banner on the webpage, is sufficient. However, Recital 134 AIA clearly indicates that the AI content itself needs a label disclosing the artificial content. Thus, deployers of an AI system that create deep fakes need to ensure that humans can observe the origin \cite{romero_moreno_generative_nodate}.

Compared to the non-trivial exceptions in Article 50(2) AIA, the exceptions for the deployers in Article 50(4) AIA are clearer. There are exceptions and limitations to this obligation for the law enforcement regime as well as for \textit{evidently artistic, creative, satirical, fictional or analogous work}. The latter exception refers to the freedom of expression and artistic creation protected by Articles 11 and EU Charter of Fundamental Rights \cite{romero_moreno_generative_nodate}.\\

The interplay between the obligations of the provider in Article 50(2) AIA and the obligations of the deployer in Article 50(4) AIA is complex and remains largely unclear. Recital 134 AIA explicitly mentions that further to the technical solutions employed by the providers of the AI system, the deployers also have to comply with the transparency obligations. It seems questionable whether double labeling adds a benefit to transparency. Standardized labeling efforts are needed \cite{gils_detailed_2024}. 

Both obligations try to regulate a technology that is, without the underlying intention of the user, neutral \cite{labuz_deep_2024,de2021distinct}. However, the overarching theme of both obligations is to ensure that even in cases where AI systems can create large amounts of synthetic images, humans are able to differentiate between synthetic and real images, see also Recital 133 AIA. This focus on the recipients of the synthetic content has already been criticized, as it does not focus on the victims of deep fakes, for example, in the case of (deep) pornographic images or videos \cite{veale_demystifying_nodate,labuz_deep_2024, meskys2020regulating}, which can cause severe harm to people. 
However, the non-compliance of providers or deployers of (commercial) AI systems with Article 50 AIA can be subject to a fine of up to €15,000,000, or, if the offender is an undertaking, up to 3\% of its total worldwide annual turnover for the preceding financial year, whichever is higher (see Article 99(4)(g) AIA). Thus, non-compliance can be very costly for a company.

\subsection{Summary of main critical legislative issues in Article 50 AI Act}
For the later discussion of the challenges, we would like to summarize our main critique relating to the legislative technique w.\,r.\,t. the transparency obligation for deep fakes and synthetic content.

First, for the definition of a deep fake in Article 3(60) AIA that is used in Article 50 AIA, it is unclear how the manipulation of content in deep fakes should be understood. This is particularly true for images, as assumptions and settings are already made by the photographer during the capture of the image, meaning that \emph{reality} is never truly represented.

Second, Article 50(2) AIA contains exceptions for assistive functions for standard editing. What can be considered as such a function? How broad is the definition of standard editing? Can the Google ``best take'' function, which is a newly developed feature, be considered as standard editing?

Third, Article 50(2) AIA also contains an exception for not substantially altering the content of an image. How broad is the definition of altering the content? What can be considered content in the sense of Article 50(2) AIA in cases of digital audio, images, and videos? For example, in the case of images, do we need to consider the pixel space or the visible space perceptible by humans? Substantive manipulations in the pixel space might not change human perception, and on the other hand, small perturbations in the pixel space can change the perception dramatically.

We address these challenges in the following section.

\section{Challenges and edge cases}
\label{sec:Challenges}
\subsection{Differentiation between legitimate AI-based image processing and deep fake generation} 
\label{sec::ImageChainDeepFake}
As discussed above, an image never depicts the objective truth. Traditional processing steps (see Section \ref{sec::post-processing-traditional}) are commonly considered to lead to authentic outcomes. Moon photos processed by Samsung smartphones (see Section \ref{sec::post-processing-ai}) do not depict reality as recorded by an image sensor but are closer to the human perception of the moon than unaltered photos. As a result, a moon photo processed in this way does not constitute a deep fake per se; as the appearance of the moon is not expected to change, the image can, in fact, be considered authentic. In summary, the photos depict the actual moon, and the AI function merely helps to overcome a technical limitation; the moon is, therefore, authentic in the resulting photos.

It is, however, conceivable that an image containing other objects besides the (post-processed) moon might have to be considered a deep fake: A viewer might assume that those other objects are depicted with the same quality as the moon, and thus falsely consider the lack of structure in these objects  authentic. A deep fake would also be created if the AI system falsely identifies an object as the moon; for example, Reddit user \emph{ibreakphotos} (see Section \ref{sec::post-processing-ai}) demonstrated Samsung's moon shot feature by taking photos of a computer screen, not the actual moon.

\subsection{Best Take}
While replacing an image of the moon with a better version may only result in deep fakes in edge cases, Google's ``best take'' seems to be an obvious example of manipulation: By exchanging a face in a group photo, the feature creates a depiction of a situation that never occurred. For such an image to be considered a deep fake, it would have to ``falsely appear to a person to be authentic or truthful'' (Article 3(60) AIA). If the photo is objectively authentic and truthful, it is not a deep fake. On one hand, in a typical group photo setting, an average viewer would probably still consider the resulting photo as authentic. The smile that is inserted existed within a couple of seconds of the remaining photo being taken. On the other hand, the ten-second time frame of the Best Take feature is sufficient for a mood change. A person might have stopped smiling while the rest of the group laughs about a joke at their expense.

As a consequence, we assume that such a group photo may well constitute a deep fake. The feature might still be ``an assistive function for standard editing'' according to Article 50(2) AIA, in which case the transparency obligation for providers of AI systems would not apply. Article 50(4) AIA, which applies to deployers of AI systems, does not include a similar exception; we still discuss whether features similar to ``best take'' can be considered an assistive function for standard editing.

One could argue that combining parts of several images goes beyond editing an image. The wording of the law, however, does not specifically refer to the editing of only one individual image. Combining information from several images has also been common, at least since the advent of digital photography. The dynamic range of a picture can be improved this way (HDR composites). In the case of panoramic shots, the individual pictures are often taken several seconds apart. This is not usually seen as  problematic. Both HDR and panoramic photos are common enough to be considered standard editing. The use of AI alone does not suffice to not consider a post-processing tool as a standard editing tool---otherwise, the exception in Article 50(2) AIA would not make sense.

On the other hand, despite being available on widely used smartphone models,  ``best take'' is marketed as a special feature (as opposed to a standard in post-processing). The difference to HDR and panoramic composites is not just the use of AI. Instead, the targeted alteration of parts of the image, and the consideration of a person's facial expression differentiate the best take feature from traditional (or ``standard'') editing.

We conclude that there is indeed a potential for the generation of deep fakes that do not fall under the exception of Article 50(2) AIA.

\subsection{Altering an image: Pixel space vs. visible space?}
\begin{figure*}
\centering
\begin{subfigure}{0.3\textwidth}
  \centering
  \includegraphics[width=0.96\columnwidth]{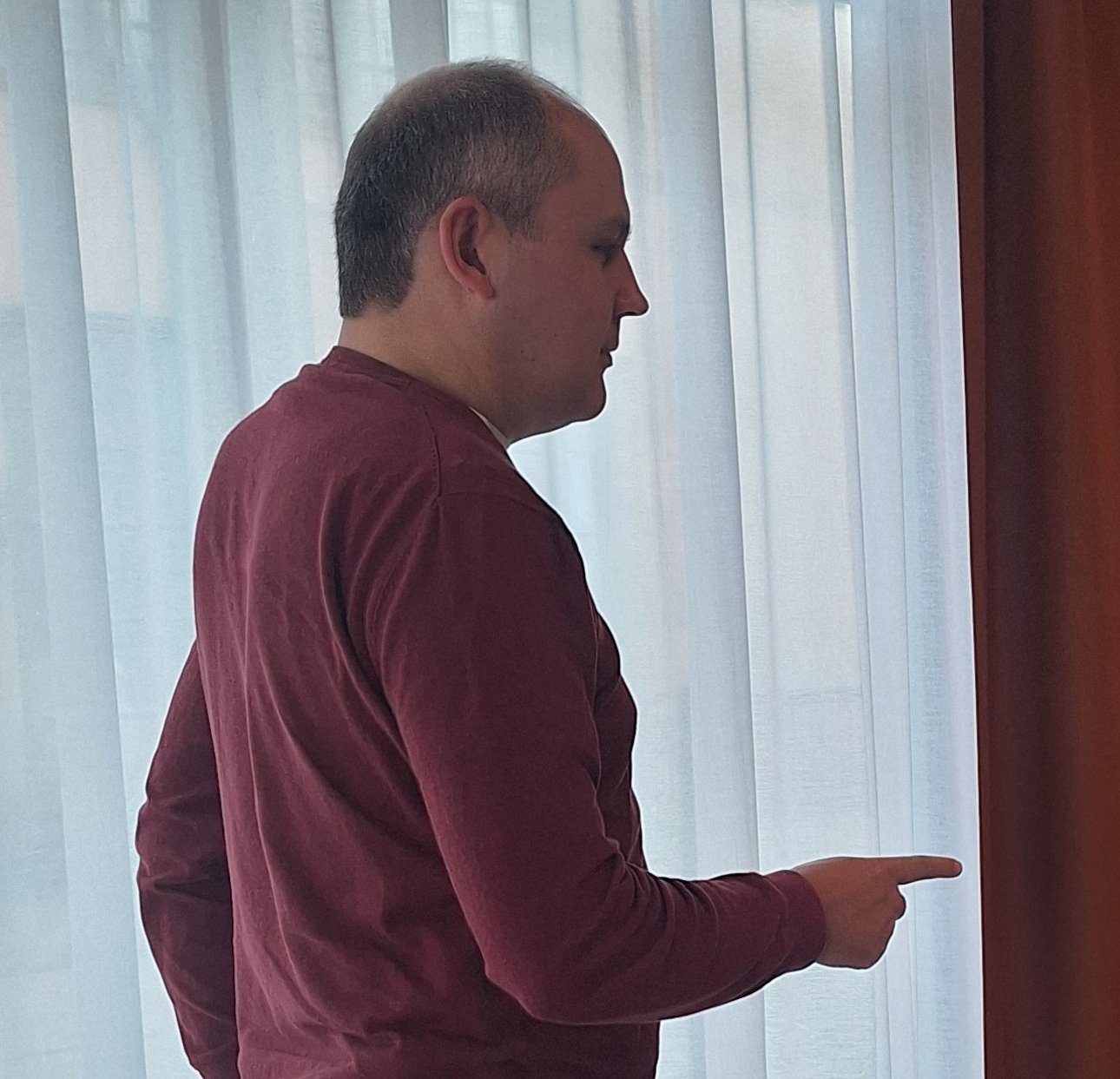}
  \caption{Original image}
  \label{fig:SorgeOriginal}
          \Description{An image of a man in front of a bright background. The man seems to be pointing at something with his right index finger.}
\end{subfigure}%
\begin{subfigure}{0.3\textwidth}
  \centering
  \includegraphics[width=0.96\columnwidth]{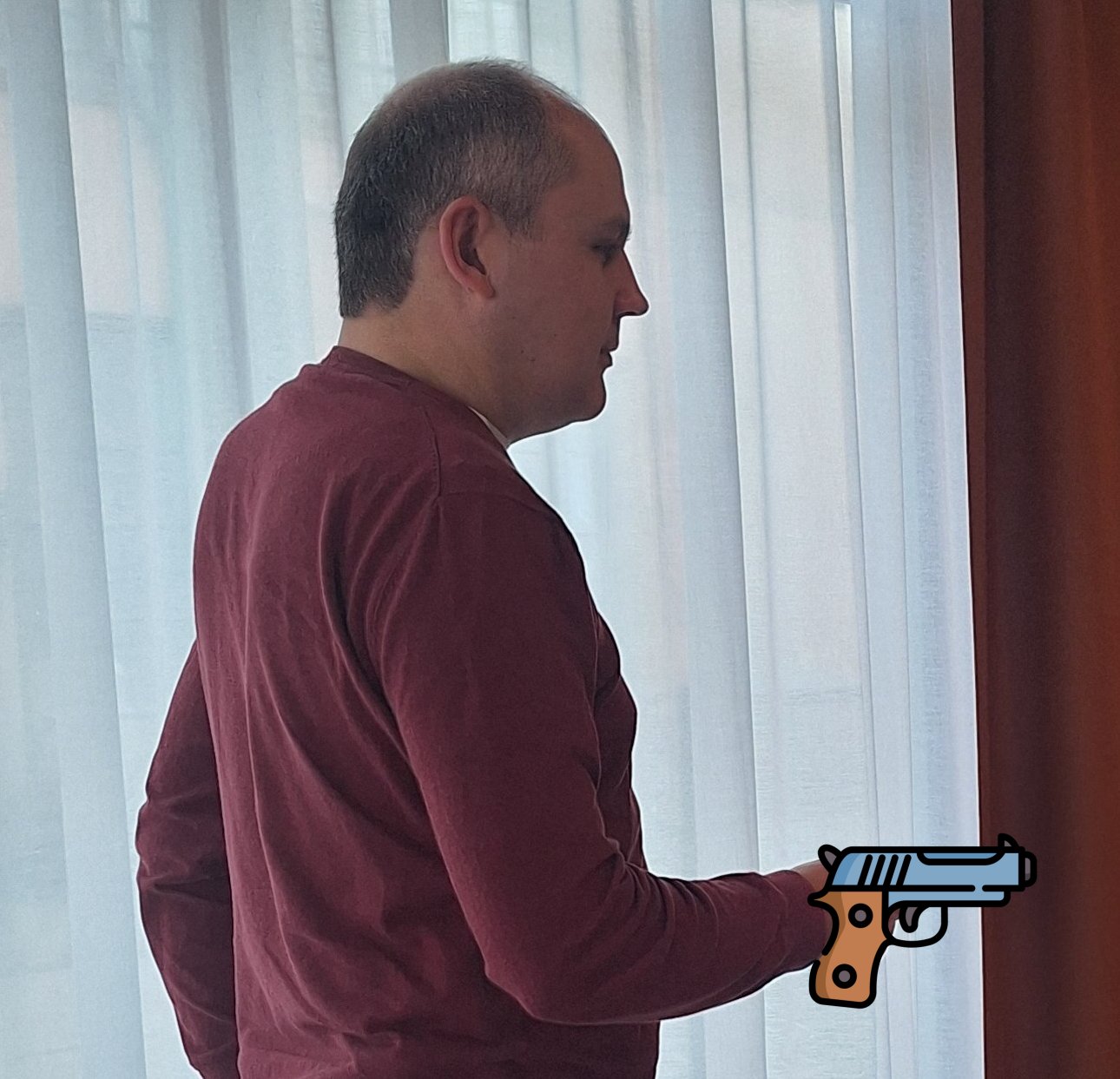}
  \caption{Version with Pistol added}
  \label{fig:SorgeIntensity}
            \Description{An image of a man in front of a bright background. The image is identical to the leftmost one, with the exception of a cartoon-like drawing of a pistol placed on top of the right hand.}
\end{subfigure}%
\begin{subfigure}{0.3\textwidth}
  \centering
  \includegraphics[width=0.96\columnwidth]{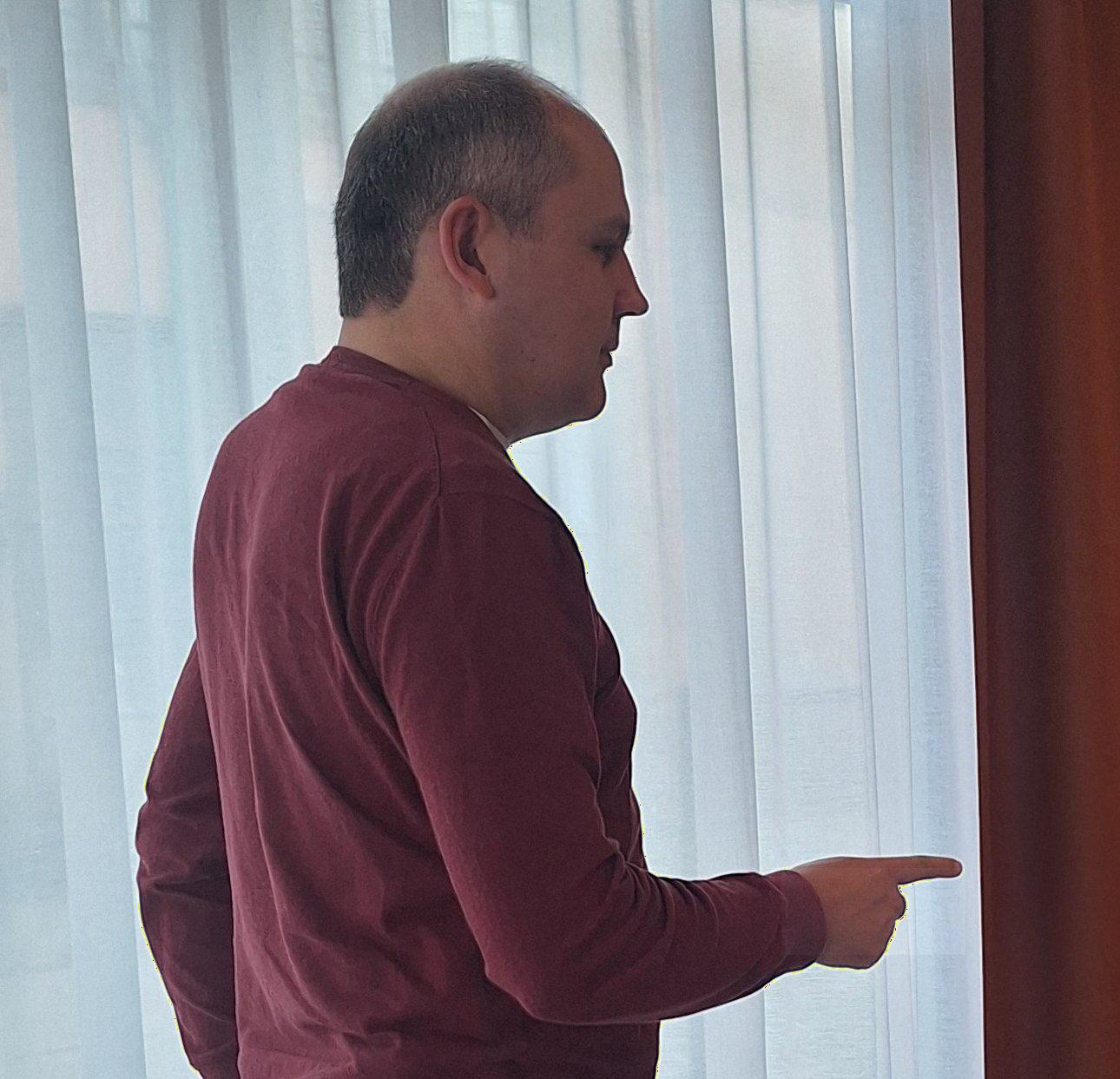}
  \caption{Version with altered intensity}
  \label{fig:SorgePistol}
            \Description{An image of a man in front of a bright background. The man seems to be pointing at something with his right index finger. The image appears identical to the leftmost one.}
\end{subfigure}
\caption{Difference in the pixel space vs. visible space. (a) Original image. (b) Added a pistol to the image. To reduce potential harm and misuse, we used a stylized pistol. (c) Image with slightly modified intensity around the hand. Figure (b) and Figure (c) have the same root mean squared difference (1.83 pixel) compared to Figure (a).  Pistol icon designed by freepik from Flaticon (flaticon.com). }
\label{fig:PixelSpace}
\end{figure*}

Besides the ``standard editing'' exception, the transparency requirement of Article 50(2) AIA does not apply to AI systems that ``do not substantially alter the input data provided by the deployer or the semantics thereof''. We consider this wording to be problematic.  While changes to the semantics are the exact reason why legislators deemed it necessary to regulate deep fakes, an exception for minor changes to the input data introduces an unnecessary backdoor. Since the exception for minor changes to the input is given equal weight alongside the semantics exception, there is little scope for interpreting the term ‘input data’ to include content-related aspects. The extent of alteration of input data is not an appropriate measure for the impact of that alteration. Fig.~\ref{fig:PixelSpace} shows an original image and two edited versions. In both modifications, the average alteration of the input data (measured by the root mean squared differences of the red, green, and blue channels for each pixel) w.r.t. the original image is identical, but the semantics are clearly different.

The exception of Article 50(2) AIA applies when either the input data (provided by the deployer) or their semantics are not substantially altered. For further discussion, we use the stylized example in Figure \ref{fig:PixelSpace}. In Figure \ref{fig:PixelSpace}(b), the semantics are substantially altered compared to Figure \ref{fig:PixelSpace}(a). But are the input data substantially altered? The manipulation in Figure \ref{fig:PixelSpace}(c) does not lead to a significant perceptible visual difference. The root mean squared differences of Figures \ref{fig:PixelSpace}(b) and Figure\ref{fig:PixelSpace}(c) compared to Figure \ref{fig:PixelSpace}(a) are identical. Thus, one can argue that even in the case of Figure \ref{fig:PixelSpace}(b), the input data are not substantially altered. Consequently, the exception applies and Figure \ref{fig:PixelSpace}(b) would fall under the scope of the exception of Article 50(2) AIA. Article 50(2) AIA would not be applicable in the stylized example.

Our example illustrates that the AI Act would benefit from a clear definition of what constitutes substantially altered content. 

\section{Recommendations}
Several issues within the AI Act and the corresponding obligations have been raised above. In this section, we want to highlight possible recommendations to mitigate these issues.

Regarding the definition of deep fakes in Article 3(60) AIA, we have demonstrated that distinguishing manipulated content from legitimate processing is challenging. However, the wording of the law gives the competent authorities and courts enough leeway to find sensible solutions for individual cases. Similarly, while examples such as the ``best take'' feature show that while a clarification of the term ``standard editing'' by the legislator might be helpful, the issue can be resolved by the judiciary. As discussed above, the term should not be applied too broadly. Even a seemingly harmless feature like ``best take'' has the potential to generate deep fakes under specific circumstances.

The transparency requirements (laid down in Article 50 AIA) on the other hand, should be clarified by the legislator. From our point of view, the exception of ``not substantially altered input data'' in Article 50(2) AIA is problematic, as alteration of data (as the term ``data'' is generally understood) is not meaningful for human perception (as we show in Figure 4). The semantics of an image can be changed in a harmful way, even if there is no substantial alteration of input data.

The exception should therefore be removed, as the other exception in the same sentence, referring to substantially altered semantics, already covers the relevant case of a changed meaning.

Furthermore, we argue that the unclear relationship between the obligations of deployers and providers in Article 50(2) AIA and Article 50(4) AIA leads to unnecessary confusion. It remains unclear whether the deployer and the provider must simultaneously label the output data. This becomes even more pressing as Article 50(2)~AIA targets deployers and synthetic content, while Article 50(4)~AIA targets providers and deep fakes. The AI Act would benefit from an explicit and congruent statement of the relationship.

As the AI Act is not likely to be amended in this way in the near future, the competent authorities and judiciary will have to deal with its shortcomings in the meantime. In addition, the European Artificial Intelligence Board, whose task is (among others) to ``collect and share technical and regulatory expertise and best practices among Member States'' (Article 66(b) AIA), may contribute to a better and harmonized understanding.

By making core issues of the regulation of deep fakes explicit, the paper at hand aims to support the competent bodies in finding pragmatic solutions in line with the legislator's intent.

\section{Conclusion and Outlook}
\label{sec:outlook}
The AI Act's definition of deep fakes, and the obligations applying to providers and deployers of AI systems that generate deep fakes, are less clear than they appear at first glance. As (digital) photos are never shown exactly as recorded by the image sensor, there has always been room for interpretation and alteration of their visual impact---even before the widespread use of AI. The delineation between legitimate processing (with the result of an image that is still considered as ``authentic'') and deep fakes (which ``would falsely appear to a person to be authentic or truthful'')  is a problematic one, and the mere use of AI tools in image processing does not suffice to cross the threshold of a deep fake. We note that regulating the use of AI systems for image manipulation is indeed crucial, as those tools are easily accessible and produce convincing images. Moreover, even legitimate AI-based image manipulation tools used without malicious intent introduce risks, as mistakes can happen in rare cases. Considering the reliance on photos (e.g., in journalism or as evidence in court), transparency requirements should not be interpreted too narrowly. A stronger focus on human oversight might be a more appropriate approach for regulating the use of AI in image editing. 

We expect discussions around deep fakes in other law areas as well. For example, there is increasing use of deep fakes in pornography, including CSAM (child sexual abuse material) and youth pornography. Prosecutors face challenges of identifying such material and have to consider its consequences for prosecution, and legislators may be urged to react to this trend as well.

With this paper, we aim to initiate an interdisciplinary discussion on the definition and regulation of deep fakes. Furthermore, we see our paper as a starting point for a dialogue between computer scientists and legal scholars on deep fake regulation. Through this interdisciplinary discussion, the regulation of deep fakes can be improved.

\section{Acknowledgments}
We would like to thank Nils Wiedemann and Roman Wetenkamp for their helpful comments on the manuscript. In addition, the authors would like to thank Thomas Wagner for providing images of the moon. Finally, we are grateful to the anonymous reviewers of the ACM Symposium on Computer Science and Law.

Kristof Meding was supported by the Carl Zeiss Foundation through the CZS Center for AI and Law. Additionally, Kristof Meding is a member of the Machine Learning Cluster of Excellence, funded by the Deutsche Forschungsgemeinschaft (DFG, German Research Foundation) under Germany’s Excellence Strategy – EXC number 2064/1 – Project number 390727645.

\bibliographystyle{ACM-Reference-Format}
\bibliography{sample-base}


\begin{thebibliography}{18}

%%% the Web address.

\ifx \showCODEN    \undefined \def \showCODEN     #1{\unskip}     \fi
\ifx \showDOI      \undefined \def \showDOI       #1{#1}\fi
\ifx \showISBNx    \undefined \def \showISBNx     #1{\unskip}     \fi
\ifx \showISBNxiii \undefined \def \showISBNxiii  #1{\unskip}     \fi
\ifx \showISSN     \undefined \def \showISSN      #1{\unskip}     \fi
\ifx \showLCCN     \undefined \def \showLCCN      #1{\unskip}     \fi
\ifx \shownote     \undefined \def \shownote      #1{#1}          \fi
\ifx \showarticletitle \undefined \def \showarticletitle #1{#1}   \fi
\ifx \showURL      \undefined \def \showURL       {\relax}        \fi
% invisible to TeX
\providecommand\bibfield[2]{#2}
\providecommand\bibinfo[2]{#2}
\providecommand\natexlab[1]{#1}
\providecommand\showeprint[2][]{arXiv:#2}

\bibitem[Becker(2024)]%
        {becker_generative_2024}
\bibfield{author}{\bibinfo{person}{Maximilian Becker}.}
  \bibinfo{year}{2024}\natexlab{}.
\newblock \showarticletitle{Generative {KI} und {Deepfakes} in der {KI}-{VO}
  {Für} eine {Positivkennzeichnung} authentischer {Inhalte}}.
\newblock \bibinfo{journal}{\emph{Computer und Recht}} \bibinfo{volume}{40},
  \bibinfo{number}{6} (\bibinfo{year}{2024}), \bibinfo{pages}{353--366}.
\newblock


\bibitem[Block(2024)]%
        {block2024critical}
\bibfield{author}{\bibinfo{person}{Martina~J Block}.}
  \bibinfo{year}{2024}\natexlab{}.
\newblock \showarticletitle{A Critical Evaluation of Deepfake Regulation
  through the AI Act in the European Union}.
\newblock \bibinfo{journal}{\emph{Journal of European Consumer and Market Law}}
  \bibinfo{volume}{13}, \bibinfo{number}{4} (\bibinfo{year}{2024}),
  \bibinfo{pages}{184--192}.
\newblock


\bibitem[De~Ruiter(2021)]%
        {de2021distinct}
\bibfield{author}{\bibinfo{person}{Adrienne De~Ruiter}.}
  \bibinfo{year}{2021}\natexlab{}.
\newblock \showarticletitle{The distinct wrong of deepfakes}.
\newblock \bibinfo{journal}{\emph{Philosophy \& Technology}}
  \bibinfo{volume}{34}, \bibinfo{number}{4} (\bibinfo{year}{2021}),
  \bibinfo{pages}{1311--1332}.
\newblock


\bibitem[Gils(2024)]%
        {gils_detailed_2024}
\bibfield{author}{\bibinfo{person}{Thomas Gils}.}
  \bibinfo{year}{2024}\natexlab{}.
\newblock \bibinfo{title}{A {Detailed} {Analysis} of {Article} 50 of the {EU}'s
  {Artificial} {Intelligence} {Act}}.
\newblock
\newblock


\bibitem[Gils et~al\mbox{.}(2024)]%
        {gils_policy_2024}
\bibfield{author}{\bibinfo{person}{Thomas Gils}, \bibinfo{person}{Frederic
  Heymans}, \bibinfo{person}{Wannes Ooms}, {and} \bibinfo{person}{Jan
  De~Bruyne}.} \bibinfo{year}{2024}\natexlab{}.
\newblock \showarticletitle{From {Policy} to {Practice}: {Prototyping} {The}
  {EU} {AI} {Act}’s {Transparency} {Requirements}}.
\newblock \bibinfo{journal}{\emph{SSRN Electronic Journal}}
  (\bibinfo{year}{2024}).
\newblock
\showISSN{1556-5068}
\newblock
\shownote{Publisher: Elsevier BV}.


\bibitem[Kristof(2024)]%
        {kristof2024online}
\bibfield{author}{\bibinfo{person}{Nicholas Kristof}.}
  \bibinfo{year}{2024}\natexlab{}.
\newblock \showarticletitle{The online degradation of women and girls that we
  meet with a shrug}.
\newblock \bibinfo{journal}{\emph{The New York Times}} (\bibinfo{year}{2024}).
\newblock


\bibitem[Lorch et~al\mbox{.}(2022)]%
        {lorch_compliance_2022}
\bibfield{author}{\bibinfo{person}{Benedikt Lorch}, \bibinfo{person}{Nicole
  Scheler}, {and} \bibinfo{person}{Christian Riess}.}
  \bibinfo{year}{2022}\natexlab{}.
\newblock \bibinfo{title}{Compliance {Challenges} in {Forensic} {Image}
  {Analysis} {Under} the {Artificial} {Intelligence} {Act}}.
\newblock
\newblock
\newblock
\shownote{arXiv:2203.00469}.


\bibitem[Meskys et~al\mbox{.}(2020)]%
        {meskys2020regulating}
\bibfield{author}{\bibinfo{person}{Edvinas Meskys}, \bibinfo{person}{Julija
  Kalpokiene}, \bibinfo{person}{Paul Jurcys}, {and} \bibinfo{person}{Aidas
  Liaudanskas}.} \bibinfo{year}{2020}\natexlab{}.
\newblock \showarticletitle{Regulating deep fakes: legal and ethical
  considerations}.
\newblock \bibinfo{journal}{\emph{Journal of Intellectual Property Law \&
  Practice}} \bibinfo{volume}{15}, \bibinfo{number}{1} (\bibinfo{year}{2020}),
  \bibinfo{pages}{24--31}.
\newblock


\bibitem[Minhaz(2020)]%
        {minhaz20android}
\bibfield{author}{\bibinfo{person}{Minhaz}.} \bibinfo{year}{2020}\natexlab{}.
\newblock \bibinfo{title}{Android Camera Subsystem -- basic image processing
  steps done at hardware level in Android Camera}.
\newblock
\newblock
\newblock
\shownote{https://blog.minhazav.dev/android-camera-subsystem-basic-image-processing-steps-done-at-hardware-level-in-android-camera/\#camera-sensor-and-raw-output}.


\bibitem[Murray(2024)]%
        {murray_generative_2024}
\bibfield{author}{\bibinfo{person}{Michael~D Murray}.}
  \bibinfo{year}{2024}\natexlab{}.
\newblock \showarticletitle{Generative Artifice: Regulation of Deepfake
  Exploitation and Deception under the First Amendment}.
\newblock \bibinfo{journal}{\emph{Available at SSRN 4872032}}
  (\bibinfo{year}{2024}).
\newblock


\bibitem[Newton(2013)]%
        {newton2013burden}
\bibfield{author}{\bibinfo{person}{Julianne Newton}.}
  \bibinfo{year}{2013}\natexlab{}.
\newblock \bibinfo{booktitle}{\emph{The Burden of Visual Truth: The Role of
  Photojournalism in Mediating Reality}}.
\newblock \bibinfo{publisher}{Taylor \& Francis}.
\newblock
\showISBNx{9781135665654}


\bibitem[Ramluckan(2024)]%
        {ramluckan_deepfakes_2024}
\bibfield{author}{\bibinfo{person}{Trishana Ramluckan}.}
  \bibinfo{year}{2024}\natexlab{}.
\newblock \showarticletitle{Deepfakes: {The} {Legal} {Implications}}.
\newblock \bibinfo{journal}{\emph{International Conference on Cyber Warfare and
  Security}} \bibinfo{volume}{19}, \bibinfo{number}{1} (\bibinfo{date}{March}
  \bibinfo{year}{2024}), \bibinfo{pages}{282--288}.
\newblock
\showISSN{2048-9889, 2048-9870}
\newblock
\shownote{Publisher: Academic Conferences International Ltd}.


\bibitem[Romero~Moreno(2024)]%
        {romero_moreno_generative_nodate}
\bibfield{author}{\bibinfo{person}{Felipe Romero~Moreno}.}
  \bibinfo{year}{2024}\natexlab{}.
\newblock \showarticletitle{Generative {AI} and deepfakes: a human rights
  approach to tackling harmful content}.
\newblock \bibinfo{journal}{\emph{International Review of Law, Computers \&
  Technology}} (\bibinfo{year}{2024}), \bibinfo{pages}{1--30}.
\newblock
\showISSN{1360-0869}


\bibitem[Shah and Kwatra(2012)]%
        {rajvi12besttake}
\bibfield{author}{\bibinfo{person}{Rajvi Shah} {and} \bibinfo{person}{Vivek
  Kwatra}.} \bibinfo{year}{2012}\natexlab{}.
\newblock \showarticletitle{All smiles: automatic photo enhancement by facial
  expression analysis}. In \bibinfo{booktitle}{\emph{Proceedings of the 9th
  European Conference on Visual Media Production}} (London, United Kingdom)
  \emph{(\bibinfo{series}{CVMP '12})}. \bibinfo{publisher}{Association for
  Computing Machinery}, \bibinfo{address}{New York, NY, USA},
  \bibinfo{pages}{1–10}.
\newblock
\showISBNx{9781450313117}
\urldef\tempurl%
\url{https://doi.org/10.1145/2414688.2414689}
\showDOI{\tempurl}


\bibitem[Szegedy(2013)]%
        {szegedy2013intriguing}
\bibfield{author}{\bibinfo{person}{C Szegedy}.}
  \bibinfo{year}{2013}\natexlab{}.
\newblock \showarticletitle{Intriguing properties of neural networks}.
\newblock \bibinfo{journal}{\emph{arXiv preprint arXiv:1312.6199}}
  (\bibinfo{year}{2013}).
\newblock


\bibitem[Veale and Borgesius(2021)]%
        {veale_demystifying_nodate}
\bibfield{author}{\bibinfo{person}{Michael Veale} {and}
  \bibinfo{person}{Frederik~Zuiderveen Borgesius}.}
  \bibinfo{year}{2021}\natexlab{}.
\newblock \showarticletitle{Demystifying the {Draft} {EU} {Artificial}
  {Intelligence} {Act}}.
\newblock  (\bibinfo{year}{2021}).
\newblock


\bibitem[Łabuz(2023)]%
        {labuz_regulating_2023}
\bibfield{author}{\bibinfo{person}{Mateusz Łabuz}.}
  \bibinfo{year}{2023}\natexlab{}.
\newblock \showarticletitle{Regulating {Deep} {Fakes} in the {Artificial}
  {Intelligence} {Act}}.
\newblock \bibinfo{journal}{\emph{Applied Cybersecurity \& Internet
  Governance}} \bibinfo{volume}{2}, \bibinfo{number}{1} (\bibinfo{date}{Oct.}
  \bibinfo{year}{2023}), \bibinfo{pages}{1--42}.
\newblock
\showISSN{2956-3119, 2956-4395}
\newblock
\shownote{Publisher: NASK National Research Institute}.


\bibitem[Łabuz(2024)]%
        {labuz_deep_2024}
\bibfield{author}{\bibinfo{person}{Mateusz Łabuz}.}
  \bibinfo{year}{2024}\natexlab{}.
\newblock \showarticletitle{Deep fakes and the Artificial Intelligence Act-An
  important signal or a missed opportunity?}
\newblock \bibinfo{journal}{\emph{Policy \& Internet}} (\bibinfo{date}{July}
  \bibinfo{year}{2024}).
\newblock
\newblock
\shownote{Publisher: Wiley}.


\end{thebibliography}

\end{document}